\documentclass[final,a4paper,twocolumn]{article}

\usepackage{amsmath}
\usepackage{amssymb}
\usepackage{graphicx}
\usepackage{graphics,epsfig,color,times}
\usepackage{multirow}

\usepackage[english]{babel}
\usepackage{url}
\usepackage[pdftex,colorlinks=false]{hyperref}
\usepackage[numbers,sort&compress]{natbib}

\usepackage[left=1.8cm, right=1.8cm,bottom=2cm,top=2.5cm]{geometry}

\graphicspath{{../graphics/}}

\usepackage[]{todonotes}

\newcommand{\fig}[1]{Fig.~\ref{#1}}

\newcommand{\tab}[1]{Tab.~\ref{#1}}

\newcommand{\eqn}[1]{Eq.~\eqref{#1}} 
\newcommand{\eqnp}[1]{\eqref{#1}} 
\newcommand{\Eqns}[2]{Equations \eqref{#1} and \eqref{#2}} 
\newcommand{\eqns}[2]{Eqs.~[\ref{#1}, \ref{#2}]} 
\renewcommand{\sec}[1]{Sect.~\ref{#1}}
\newcommand{\vid}[1]{Video~\ref{vid:#1}}
\newcommand{\CharOver}[2]{\ensuremath{\overset{_{#1}}{#2}}}

\newcommand{\ie}{i.\,e.~}
\newcommand{\cf}{c.\,f.~}
\newcommand{\eg}{e.\,g.~}

\newcommand{\Real}{\ensuremath{\mathbb R}}        
\newcommand{\T}{\ensuremath{\top}}                

\title{Self-organized control for musculoskeletal robots}
\author{Ralf Der$^1$ and Georg Martius$^{2}$\\[.5em]
{\normalsize $^1$\,Max Planck Institute for Mathematics in the Science, Leipzig, Germany}\\
{\normalsize $^2$\,IST Austria, Klosterneuburg, Austria}\\
 {\tt \small <ralfder@ist.ac.at>, <gmartius@ist.ac.at>}
}

\begin{document}
\maketitle

\begin{abstract}

With the accelerated development of robot technologies, optimal control becomes one of the central themes of
research. In traditional approaches, the controller, by its internal functionality, finds appropriate actions on the basis of the history of
sensor values, guided by the goals, intentions, objectives, learning schemes, and so on planted into it.
The idea is that the controller controls the world---the body plus its environment---as reliably as possible.
However, in elastically actuated robots this approach faces severe difficulties.
This paper advocates for a
new paradigm of self-organized control. The paper presents a  solution  with a controller
that is devoid of any  functionalities of its own, given by a fixed, explicit and context-free function of the recent history of
the sensor values.
When applying  this controller to a muscle-tendon driven arm-shoulder system from the Myorobotics toolkit, we
observe a vast variety of self-organized behavior patterns: when left alone, the arm realizes pseudo-random sequences
of different poses but one can also manipulate the system into definite motion patterns.
But most interestingly, after attaching an object, the controller gets in
a functional resonance with the object's internal dynamics:
when given a half-filled bottle, the system spontaneously starts shaking the bottle so that maximum response from
the dynamics of the water is being generated.
After attaching a pendulum to the arm, the controller
drives the pendulum into a circular mode. In this  way, the robot discovers dynamical affordances of
objects its body is interacting with.
We also discuss perspectives for using this controller paradigm for intention driven behavior generation.

\footnotesize
Keywords: self-organization, robot control, musculoskeletal, tendon-driven, learning, anthropomimetic, self-exploration
 \end{abstract}

\section{Introduction}
Control is a ubiquitous theme of life and technology. When reaching for a cup of coffee or walking through the mountains,
our neural systems control all movements with great ease,
despite the great uncertainty involved in controlling the muscles,
the complexity of the task
and many other factors.
That this simplicity is an illusion is seen as soon as trying to program a robot for doing a task.
While the complexity of programming stands as a challenge for decades, in recent times considerable progress
has been achieved by new materials~\citep{Kim2013:Softrobotics},
 powerful actuators~\citep{RaibertEtAl2008:Bigdog},
 the improved theory of control~\citep{Siciliano2009:Robotics},
 but in particular by the tremendous increase in
 computational power that allows modeling and physically realistic simulations of very complex systems
to improve planning and control~\citep{MordatchTP12:CIO,ErezTodorov2013:MPC-humanoid,PosaTedrake2014:Traj-direct}
and even allows to simulate large controlled muscular body systems~\citep{YamaneNakamura2011:modelinghuman},
or find new perspectives for artificial evolution~\citep{bongard2015using} by exploiting super computer power.
However, any approach setting on computer power finds its boundaries promptly by the combinatorial complexity of robots in the environment.
The DARPA challenge presents numerous examples of progress but also reveals  a realm of failures of these systems
even under remote control.
Also there are a variety of new control paradigms around, best demonstrated by the amazing locomotion abilities  of the
Boston dynamics robots, like BigDog, PETMAN and others.
These are ingeniously  engineered systems for realizing a specific set of tasks with their highly specialized bodies.

The present  hype in ``classical'' robotics  evokes the early times of  AI
with its firm believe that the behavior of robots can be prescribed on
a formal level by formulating a set of rules to be realized by the machine.
In practical applications, this approach proved very soon to fail due to the complexity of both modeling and
controlling real world scenarios.
A rethinking began more than two decades ago,
with Rodney Brooks statement ``the world is its own best model'',
prompting a drastic change of
paradigms in the control of autonomous
robots~\citep{brooks85:layered,brooks91:representation}. The so-called embodied AI
recognizes the body as an equal partner in the control process.
The exploitation of the specific properties of the body, sometimes
called
{\it morphological computation}~\citep{Hauser2012,Pfeifer2009,Paul:2004:MC}
has subsequently become an active field of research with many impressive results,
see~\cite{PfeiferBongar2006:BodyShapesThink,pfeifer99:understIntel}
opening new perspectives for both robot control and our  understanding human sensorimotor intelligence~\citep{pfeifer2012}.

To date, the two branches of robotics---the classical AI versus the embodied approach---coexist,
each one having its realm of relevance. The embodied approach seems to be favored in systems with strong
physical effects, like soft robotic systems or elastically actuated robots, where the engineering approaches run into severe difficulties.
The limitations of present day engineering approaches to human like structures is best seen when considering
muscle-tendon driven (MTD) systems where an important line of research was shaped by  EU projects leading from CRONOS, to
ECCEROBOT to MYOROBOTICS, but also by Japanese projects creating the highly sophisticated Kenshiro robot~\cite{KozukiEtAl2012:ShoulderThoraxOfKenshiro,NakanishiEtAl2013:MusculoskeletalHumanoids}.
While excellent work has been done in planning, constructing, and eventually building these robots~\citep{wittmeier2013a},
the control of these systems~\citep{potkonjak2011anthropomimetic,potkonjak2011puller,NakanishiEtAl2013:MusculoskeletalHumanoids} is still in its infancy.
Problems stem from the fact that every movement leads to a whole body answer and the effect of an action
 has much different effects depending on the state of the system.
One option is to model the system in order to map it to a conventional rigid joint system, with workspace analysis~\citep{LauOetomoHalgamuge2013:staticworkspaceMTD} and actuation effects computed by a suitable cable routing matrix approach~\citep{LauOetomoHalgamuge2013:ModelForCableDrivenManipulators}.
However, measuring joint angles precisely is a problem in practice and also to account for
 tendon/muscle elasticities causes sever inaccuracies especially when it comes to dynamics.

The other option is to use learning to achieve a new behavior, \eg with reinforcement learning.
For this a suitable exploration has to be
 implemented, because a random unstructured search is not feasible.
We argue that a new control paradigm based on self-organization can bootstrap the control problem
 by providing basic dynamic behaviors in a very short time, which can be later combined to solve tasks by known learning methods.
The idea of self-organizing behavior is quite different from classical control
 because there is {\bf no} target signal or any other specific input to the system.
The only source of information are the sensor values itself, containing the responses of the world to action
  on which a suitable generic drive has to be implemented.
In this work we propose a generic drive that increases velocity correlation between the degrees of freedom.
The concrete new behavior emerges from old behavior in a dynamical
 process where the classical roles of the controller and the controlled are seemingly inverted.
This paper demonstrates that a controller, only given in terms of responses
 of the world without a predefined goal, leads to self-organized control producing a wealth of meaningful behavior
in embodied robotic systems.
The paper also  outlines some of the most prominent consequences.
Our approach somewhat implements the idea of \citet{PfeiferBongar2006:BodyShapesThink} that ``the
body shapes the way we think'' directly into a low-level control framework,
letting the body become actual creator of the robot's basic way of acting and eventually thinking.

There are also other approaches for creating self-driven behavior from generic drives.
There is predictive information maximization~\citep{MartiusDerAy2013} which creates less coordinated behaviors than
 the presented approach, but is similar in the setup.
Active goal exploration~\citep{BaranesOudeyer2013:ActiveGoalExploration}
 can be considered as an implementation for high-dimensional systems of the long
 suggested maximizing of learning progress~\citep{Schmidhuber91:CuriousControl}.
It was not yet applied to soft robots, requires a much longer time for new behaviors and is not interactive.
However, it results in a complex internal model of the robot which we do not train here.

For our experiments we have chosen an anthropomimetic musculoskeletal arm-shoulder platform that is actuated by
 tendon-spring elements. This robot has many degrees of freedom (10) and is elastically actuated which makes it
 hard to be classically controlled.
We demonstrate that our control approach creates an meta-system---formed by controller, body, and environment---with a rich variety of all kinds of attractors.
These can be deliberately switched  by manipulative disturbances, creating an attractor meta-dynamics~\citep{gros2014attractor}.
We use the same controller without any changes for all experiments.
There is no reward or specific goal, just the responses of the world (sensor values)
 which guide the behavior by the generic drive to increase velocity correlations between the joints.
This is seen by the ``willingness'' of the meta-system  to follow and repeat manually imposed motion patterns.
More interestingly, the meta-system
may become a resonator which is excited by the self-amplification of latent motion patterns of its physical subsystems.
For instance, when suspending a weight from the tip of the arm,
the meta-system is  piloted  by this pendulum into a resonant state where the pendulum weight describes a circular motion pattern.
Different from a chain carousel
which is driven by an outside torque, this motion pattern emerges only by the
self-amplification of the tiny but systematic forces exerted by the pendulum on the muscles of the arm. In another setting,
when given a bottle half-filled with water, the robot starts shaking the bottle in
a definite manner, driven by the dynamics of the water.

These and many more behaviors can be observed in a single run,  without stopping the system or
manipulating the controller in any way, and is a direct consequence of using the physical responses
 as the only source of information. This work is a further development of our previous work~\citep{DerMartius2015:DEP}, putting the approach on a more
 solid theoretical basis and demonstrating its usefulness with a really challenging physical system.
 We also discuss perspectives for using this controller paradigm for deliberate behavior generation.
This is the decisive next step that will leverage  the approach to a versatile tool for behavior generation of soft robots.

\section{Self-organized control for soft robots}
The controller, we propose, is a function that receives
at time $t$ a vector of  sensor values $x_t \in \Real^n$ and
sends a vector of motor values $y_t\in \Real^m$.
As we aim at self-organization of control, we have to define the control signals in a self-consistent way
on the basis of the history of sensor signals alone.
Let us introduce  $x'_t=x_{t+\theta}$, the vector of the  future sensor values, where $\theta $ is a time lag with $\theta=1$
 in the derivations given below (time is measured in discrete update-steps, here up to 100/sec).

In order to have non-trivial behavior the sensorimotor loop needs to be mildly destabilized,
driving it  into self-excited behavior modes.
In this sensorimotor loop the controller regulates the energy feed-in.
Destabilization can be achieved by increasing the overall feedback strength.
However, to keep this destabilization in bounds, we need a conservative element which is formulated by the following
self-consistency requirement.
We postulate the existence of a forward model given by the (state dependent) matrix $A$ so that
\begin{equation}
  x'_t=A_t y_t +\xi_t
  \label{eqn:forward}
\end{equation}
where $\xi$ is the modeling error. This describes the physical dynamics over one time step.
Introducing $M$ which is the inverse or pseudoinverse of $A$ we
require $y$ to be a function of the future sensor values $x'$,
 \begin{equation}
  y_t \CharOver{!}{=} M_t  x_t'
\label{eqn:dotxdotz}
 \end{equation}
This is the conservative element as it binds the control of the system and hence the energy feed-in
 to the real physical situation.
So, \Eqns{eqn:forward}{eqn:dotxdotz} display the essential idea of our approach to make
 motor signal compliant with the world dynamics.
In a sense, \eqn{eqn:dotxdotz} means that the world's responses, represented  by $x'$,  signals the controller what to do.
But of course the world (\ie the future sensor values $x_t'$) is also controlled by the controller
 through the actions $y$ \eqnp{eqn:forward}.

The choice of the sensor-to-motor mapping $M$ will be discussed below, see \sec{sec:roleofM}.
However, it only has to capture the coarse causal relationship between
 sensor and motors and even in the case of the tendon driven robot a one-to-one mapping is used ($M$ is the identity matrix)
 because each motor position is related to the tendon length on a one-to-one fashion.
In this way  \eqn{eqn:dotxdotz} boils down to $y_t \CharOver{!}{=} x_t'$.

However, we cannot use \eqn{eqn:dotxdotz} directly for generating the control signal $y$ as it contains the future.
So, we must find a model for relating the future sensor signals $x'_t$ to their past, \ie  $x_t, x_{t-1}, \ldots$.
In other words, we need a time series predictor for the sensor dynamics.
This may be obtained in many different ways, however,
 we propose one that is particularly simple and
 creates a general drive to increase velocity correlation between the degrees of freedom,
 see also Appendix \sec{sec:Appcontroller}.
It consists of a time varying linear predictor for the derivatives of the sensor values
 in the form $\dot x_t' = \bar L \dot x_t$.
The matrix $\bar L$ is given by an exponential weighted time average of the past velocity correlations as
\begin{equation}
  \bar L(t)= 1/Z \sum_{s=\theta}^{t-1} \rho^{t-1-s} \dot x_{t-s}' \dot x_{t-s}^\top/\| \dot x_{t-s} \|^{-2},\label{eqn:barL}
\end{equation}
where $\rho<1$ and $Z$ is a normalization constant.
The important part is the term $\dot x_{t-s}' \dot x_{t-s}^\top$ measuring the velocity correlations between sensor vectors with a given time lag.
The predictor $\bar L$ is then used in the following controller
\begin{equation}
   y_t=g(C_t x_t)
   \label{eqn:controls}
\end{equation}
where  $C=  M\bar L $ (all quantities at time t).
The function  $g: \Real^m \rightarrow \Real^m$ is a squashing function so that motor values are kept in bounds.
 In the applications we use the $\tanh$ function applied element-wise.
Note that the matrix $C$ can also be obtained from an update rule, see \sec{sec:Appupdate},
 which differs from earlier work~\citep{DerMartius2015:DEP}
 by the normalization factor $\|\dot x\|^{-2}$ in \eqn{eqn:barL}.
In the experiments this leads to a more continuous activity in the behaviors avoiding long pauses of inactivity.

The self-consistency principle formulated by \eqn{eqn:forward}--{eqn:controls} is not yet operational.
There is a global attractor given by $y_t=0$ for all $t$ (note that the squashing function $g$ is antisymmetric),
 \ie in a position controll setting all actuators are in their central position given by $y=0$.
In order to destabilize this attractor, we use that the sensorimotor system is a feedback loop with the
controller regulating the energy feed-in.
For increasing the overall feed-back strength,
we replace the matrix elements of $C$ in \eqn{eqn:controls}
as $C_{ij} \leftarrow \kappa C_{ij}/\left({\|C_i\|} + \lambda\right)$,
 with $\|C_i\|$ denoting the norm of row $i$  and
 $\lambda\ll 1$ is a regularization for keeping the normalization factor in bounds.
$\kappa$ may be considered as a kind of ``character'' parameter as it  defines essentially the
amplification factor of the loop. Using  $\kappa>\kappa_c$, where $\kappa_c\approx 1$,
it determines the amplitude of the emerging motion patterns.
A second such ``character'' parameter is given by the  time scale $\rho$ for the history  of the sensor values included in $C$, see the Appendix for details.
The implementation of the controller is thus simple and boils down to
 just two equations \eqn{eqn:barL} and \eqn{eqn:controls} with the normalization of $C$ described above.

This controller paradigm  differs from usual paradigms in many ways.
There is no specific goal, no target signal, no learning, no biasing, no sampling,
 and no preprocessing or preconditioning.
There are just the two ``character'' parameters $\rho$ and $\kappa$.
All the controller does is to push incoming sensor vectors into the history stack and use the latter for evaluating $C$.
Why is such a simple controller able to generate the different behaviors by itself?
When considering $\bar L$ we see that it basically contains the \emph{velocity correlations}
 of subsequent sensor vectors in the recent past such that the system is driven towards patterns with high velocity correlations in the sensor values.
Any initial movement or perturbation sets the germ for a small movement which is then amplified due to the positive feedback strength set by $\kappa$, if sufficiently large.
However, an unbounded growth of the motor values is not possibly due to the non-linear controller ($g$ function)
 which leads to a confinement.
A solution to the requirement of maintaining high velocity correlations and a positive feedback loop is to
 enter an oscillation.
The orbit of the limit cycle is however strongly influenced by the velocity correlation
 which are determined by the embodiment effects.
Behaviors become stationary if they fulfill the \emph{self-consistency} where $\bar L$ given by \eqn{eqn:barL}
 produces with the behavior which sustains $L$.
This is the case for harmonic oscillations where $\bar L$ become more or less a constant rotation matrix.
However, also a limit cycle in the $L$ dynamics is possible leading to a periodic change in the behavior.
Until a stationary behavior is found the system wanders in a long transient through many different
 oscillatory behaviors, in some cases this process may not end because no self-consistent behavior is found.
A deeper mathematical analysis is left for future work.

The essential feature is the irreducible unity of the controller and the controlled
 in an adaptive self-referential dynamical system -- a meta system.
The idea to consider the sensorimotor loop as a closed meta-system is also central to a body of work,
see for instance~\cite{SandorGros2015:sensorimotor,gros2015complex,toutounji2014behavior,toutounji2016autonomous}.

\section{Experiments}

\begin{figure}
  \centering
  \includegraphics[width=\linewidth]{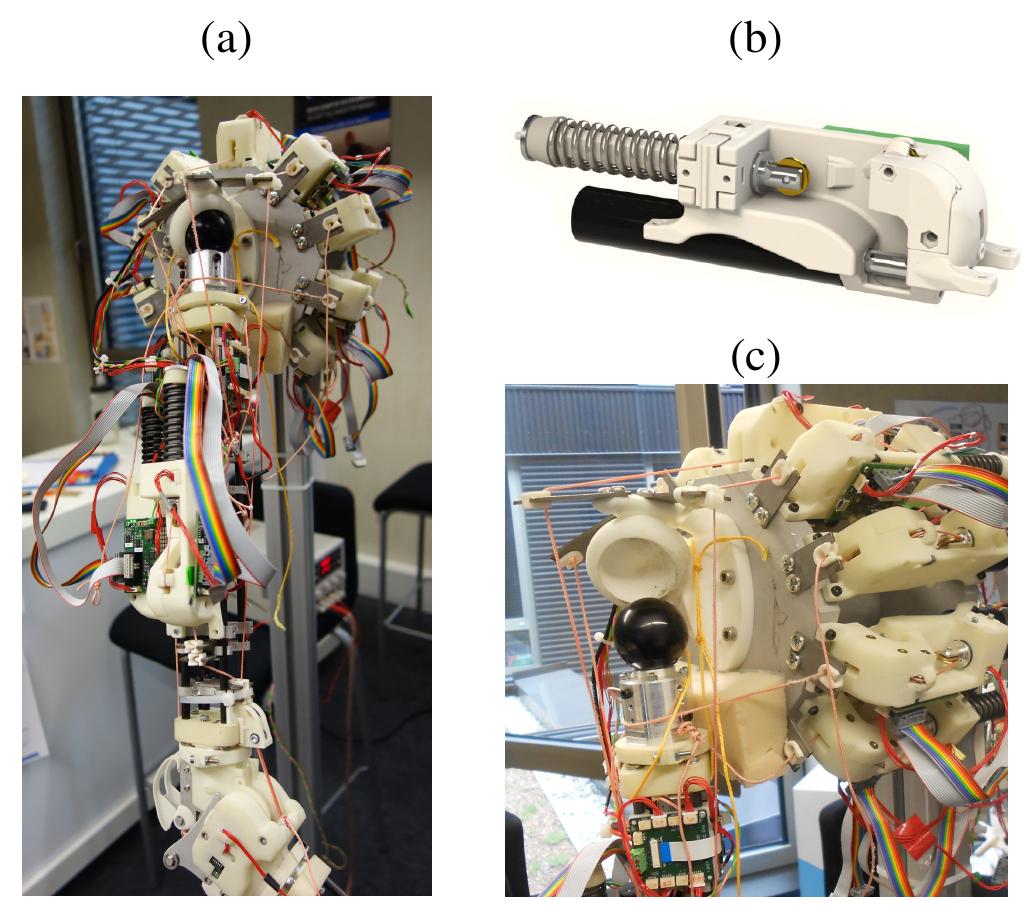}\\[-.5em]
  \caption{{\bf Myorobotic arm (a), a single muscle element (b), and a dislocated shoulder (c).}
    The dislocation happens wickedly as soon as the tendons are getting slack.
    \label{fig:arm}
  }
\end{figure}

The above defined controller was used in the experiments with a tendon driven arm-shoulder system from the Myorobotics toolkit~\citep{Marques2013}, see \fig{fig:arm}.
The system has 11 artificial muscles, 8 in the shoulder and 2 in the elbow and one affecting both. However
 two of the shoulder muscles where disconnected.
The muscles are composed of a motor winding up a tendon connected to a spring, see \fig{fig:arm}(b).
The length of a tendon $l$ is given by the motor encoders
and the spring force is translated
into  a length  $s$  in the interval $[-\alpha,1-\alpha]$ where $\alpha$ defines pretension (here $\alpha=0.1$).
The length of the tendons is normalized to $l\in[-1,1]$.
We define the sensor values as
\begin{equation}
  x_i = l_i + \beta s_i \label{eqn:sensorvalues}
\end{equation}
where $\beta$ regulates the integration of the spring-length. In the experiments,
$\beta$  was simply set to 1 without further tuning. It is expected that this choice is not critical.

\subsection{Peculiarities of muscle-tendon driven systems}
\label{sec:MTD}
There are a number of features which make the muscle-tendon driven (MTD) systems different from classical robots
having revolute joints with direct motor control, \ie the motor positions directly translate into joint angles and into
poses.
The most obvious effect is seen  when tendons are getting slack so that contact with the physical
state of the arm is lost altogether.
This has to be avoided by keeping a permanent tension on the tendons, which poses another problem:
The tension can only be achieved by tightening each tendon up against all the others,
each individual tension being reported by the spring length. This means that (i) there are infinitely many combinations of tension
forces for a single arm pose and (ii) that the action of a single motor will be reflected in a change of
spring length of all other muscles. In other words, actuating a single muscle is reflected by a pattern of sensory
stimulation--- a whole-body answer.

Furthermore, the combination of friction effects and muscle-pose ambiguity leads to a strong  hysteresis effect.
After driving the arm by a sequence of motor commands from pose A to pose B
one ends up in a different pose and muscle configuration than $A$ after moving back by reversing the motor commands.
In general, this makes the translation of a kinematic trajectory for the arm into motor programs extremely difficult, even
more so if there are loads and high velocities involved. Also, the classical approach of
learning a model by motor babbling becomes illusory.
In a sense, these problems dissolve partially
 when using the system's physical responses for finding the control signals.

\subsection{Self-regulation, manipulability, and goal orientation}
We conducted several experiments listed in \tab{tab:experiments} which demonstrate the essential features of the control scheme.
All experiments are done with the same controller with the same initialization ($C=0$) so that it is
only the physical situation that differs between the experiments.
We will keep this paper mainly on the level of phenomena, providing a deeper analysis in a later paper.
We recommend consulting the videos for better understanding.
\newcounter{video}
\refstepcounter{video}\label{vid:demo}
\refstepcounter{video}\label{vid:handshake}
\refstepcounter{video}\label{vid:bottle:swing}
\refstepcounter{video}\label{vid:bottle:shaking:horizontal:white}
\refstepcounter{video}\label{vid:bottle:shaking:horizontal}
\refstepcounter{video}\label{vid:bottle:shaking:vertical}
\refstepcounter{video}\label{vid:free:moving}

\begin{table*}
  \caption{
    {\bf Experiments.} The videos can be watched at \href{http://playfulmachines.com/MyoArm-1}{{\tt http://playfulmachines.com/MyoArm-1}}. An overview is given by a cut version in \href{http://playfulmachines.com/Videos/MyoRobotics/MyoArm-DEP_small.mp4}{\vid{demo}}. \label{tab:experiments}}
  \centering\small
  \begin{tabular}{lp{.65\linewidth}p{1.5em}l}
    Title        & Description&Sect.&Vid.\\
    \hline
    Handshake    &Human robot interaction by manually imposing a periodic movement &\ref{sec:mani}&
    \href{http://playfulmachines.com/Videos/MyoRobotics/MyoArm-handshake-small.mp4}{\vid{handshake}}\\
    Bottle swing&Suspending a weight from the tip of the arm: excitation of a circular pendulum  mode&\ref{sec:pendulum} &
    \href{http://playfulmachines.com/Videos/MyoRobotics/MyoArm-bottle-swing_small.mp4}{\vid{bottle:swing}}\\
    Shaking horizontal   &A quarter filled bottle is horizontally attached to the tip of the arm: shaking of the bottle  mainly along its axis&\ref{sec:bottle:shaker}&
    \href{http://playfulmachines.com/Videos/MyoRobotics/MyoArm-whitebottleshake-long_small.mp4}{\vid{bottle:shaking:horizontal:white}}\\
    Shaking hori. II &Second example of the horizontally bottle case with more liquid&\ref{sec:bottle:shaker}&
    \href{http://playfulmachines.com/Videos/MyoRobotics/MyoArm-bartender-hor_small.mp4}{\vid{bottle:shaking:horizontal}}\\
    Shaking vertical&The same as above but with vertical attachment&\ref{sec:bottle:shaker}&
    \href{http://playfulmachines.com/Videos/MyoRobotics/MyoArm-bartender-vert_small.mp4}{\vid{bottle:shaking:vertical}}\\
    Free            &No external forces applied: pseudo-random sequences of reaching-type behavior&\ref{sec:reaching}&
    \href{http://playfulmachines.com/Videos/MyoRobotics/MyoArm-widemoves_small.mp4}{\vid{free:moving}}\\
  \end{tabular}
\end{table*}

\subsubsection{Self-regulated working regime}\label{sec:regime}
Before presenting  the experiments in more detail, let us take a look
 at the sensorimotor coupling that is created by this controller.
One of the crucial features is the self-regulation into a working regime where the
 tendons are kept under tension even in very rapid motions with notable loads.
This is very important as it guarantees the signals from the controller to be executed in a definite way.
As a result, in all experiments we never had to face a shoulder dislocation, see~\fig{fig:arm}(c),
 which may happen promptly if tendons are getting loose.
This is the more astonishing as this sensible working regime emerges  without any additional
tuning or calibrating~\citep{wittmeier2012calibration} the system.
For that, the particular sensor configuration \eqn{eqn:sensorvalues} seems to be important, but we did not
 study it systematically yet and expect other configurations to work as well.

\subsubsection{Manipulability}\label{sec:mani}
The dominance of the sensor responses in the control framework makes the
 controlled system manipulable by externally applied forces.
The point is that any additional forces applied to the arm segments
 change the sensor values via the changing spring tension. This effect integrates manipulative influences---like a physical robot
human interaction---into the sensor values and thereby, via the controller matrix $C$, in the behavior generation.
For instance,
 the arm can always be stopped by applying a force by hand.
The reason is not  that the motors are too weak.
Instead, $\dot x=0$ is a fixed point of the dynamics the meta-system to which it relaxes
if the mechanical degrees of freedom are frozen manually~\footnote{This effect involves  the normalization
factors and fades away once the regularization comes into play. After that, the system tries to move to the global attractor $\dot x =x =0$.}.

Moreover, the system can be entrained by manual interaction into specific behaviors.
We demonstrate this in the handshake experiment, see \vid{handshake},
 where the user is trying to  move the arm in a periodic pattern.
Besides the possibility to train a robot in this way, the most interesting point is
the subjective feeling that comes about when interacting with the robot.
In the beginning of such an interplay, the robot seems to have a will of its own as
it resists the motions the user is trying to impose. But after a short time the robot follows the human more and more and eventually
is able (and ``willing'') to uphold the imposed motion by itself.
Otherwise,  depending also on the human partner, the meta-system of robot and human may ``negotiate'' a joint motion pattern
which might be  left if the human quits the loop.
This can be understood by realizing that many periodic patterns are consistent with
 \eqns{eqn:controls}{eqn:App:barLm}. If the imposed patterns matches one of them then the robot is controlling this pattern by itself.
In fact, in the experiments, one can well observe that a ``compliant'' human is intrigued to follow the
system as much as its own intentions, ending up in  an orchestrated human-machine dynamical pattern.

The training of a robot by directly imposing motions is not new. The common approaches generate a
kinematic trajectory which is afterwards translated into the motor commands by well known engineering methods.
We argue that this method will run into severe difficulties due the peculiarities of our MTD system as discussed in \sec{sec:MTD}.
On the contrary, in our approach we have an entrainment effect in the meta-system (body plus controller)
which does not need kinematics and dynamical modeling at all.

\subsubsection{Perspectives for goal oriented behavior}
\label{sec:goaloriented}

Another aspect is given by the challenging task of deliberate control, like
realizing a given behavior. Let us consider the perspectives of our  approach for that task.
The  idea is based on the observation that the  meta-system (the physical system with the controller regulating the energy feed-in)
has totally  different physical properties than the ``bare'' or passive system (motor power switched off).
In particular, with  $\kappa\approx \kappa_c$, self-amplification is subdued so that
the meta-system goes into a plastic aggregation state.
In this specific regime, patterns are present as mere  potentialities waiting to be excited. It is easy to conceive that
through a second controller---let us call it the meta-controller---acting on the meta-system,
it is
more reliable to drive the meta-system
toward those latent behaviors than  by acting on the ``bare'' physical system.
In fact, if we are able to influence the meta-system by hand, why not
by just superimposing additional motor signals on the self-regulated meta-system.
The use of the approach is encouraged by the mentioned  ability of the meta-system to uphold a resilient working regime (see above)
even under extreme external perturbations, preventing, for instance, shoulder dislocations.

\subsection{Emerging modes}
As mentioned already above, with the transition operator $\bar L$, the meta-system is particularly akin to
periodic motions, \ie there is a plenitude of  latent limit cycle attractors which, metaphorically speaking,
wait for their excitation. The selection of a specific attractor
 is realized by the self-amplification of a dynamical seed, generically  provided by a
physical subsystem (of the world) with a time coherent internal dynamics.
In other words, the subsystem---by its internal dynamics---is piloting the
meta-system  into a resonant state, \ie a whole-system mode with defined frequency.
This salient phenomenon is a direct consequence of
giving the world the leading role in behavior generation.
In the following experiments, we additionally feed in delayed sensor values which enhances the affinity for periodic behaviors, see \sec{sec:periodicity}.

\subsubsection{Self-excited pendulum modes}
\label{sec:pendulum}
In a first experiment, we suspend a weight  (the bottle) from the tip of the arm.
The world in this experiment consists of the arm together with a physical subsystem---this pendulum---with a complex internal dynamics.
With the pivot point at rest, the pendulum  may realize ellipsoidal or even circular motion patterns with fixed frequency.
With  a driven pivot the pendulum is able of chaotic motions under certain trajectories of
 the pivot point.
In any case, the motions of the weight exert small forces on the arm which change the spring tensions and thereby the sensor values.
While being tiny, these reactions are systematic so that they may accumulate  in $C_{ij}$. The emerging pathways in the feedback loop
sensitize  the
meta-system to potential pendulum modes, readying itself for their
self-amplification.

\begin{figure}
  \centering
  \includegraphics[width=\linewidth]{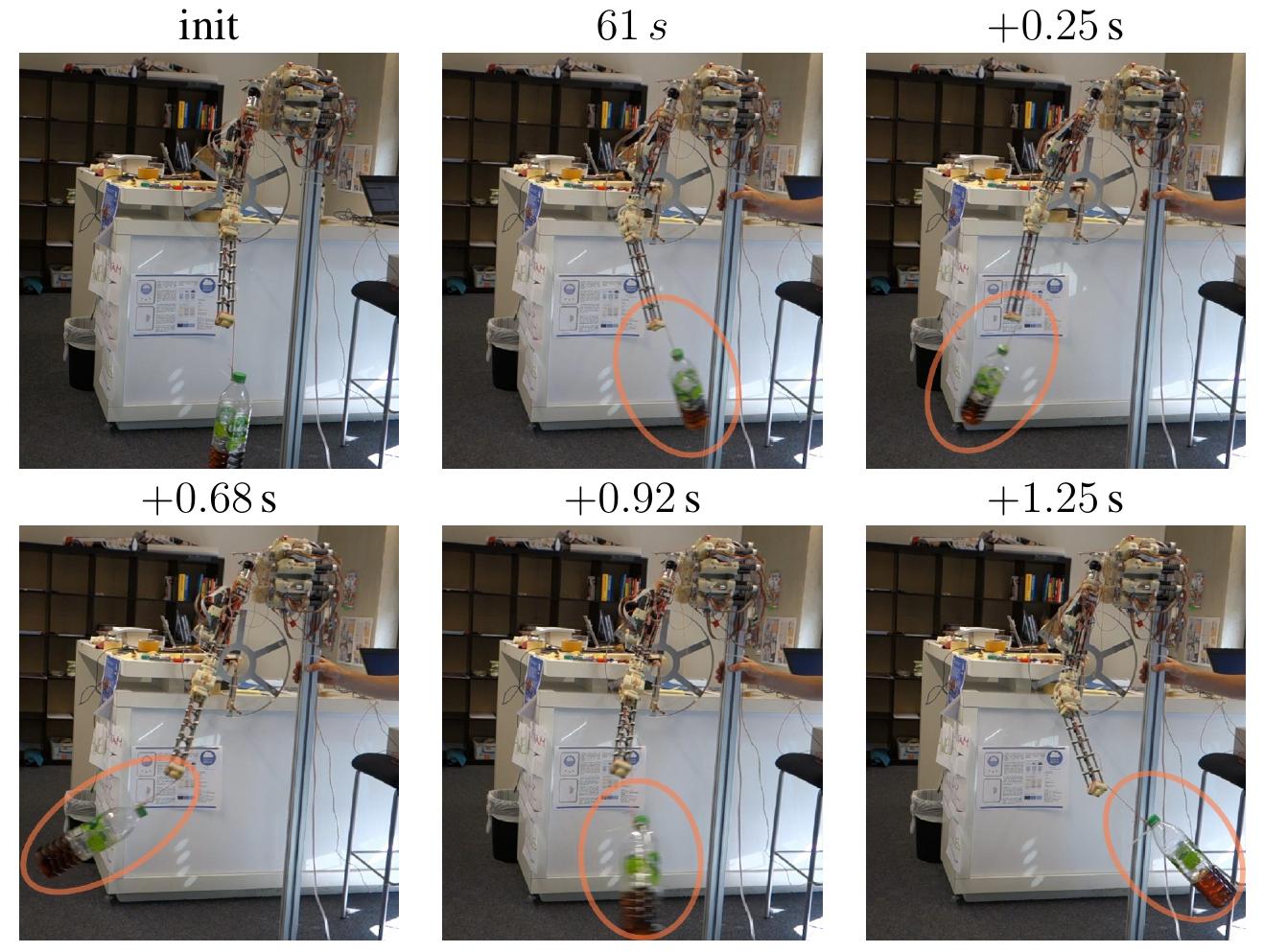}\\[-.5em]
  \caption{{\bf Swinging the bottle.} Still images of \vid{bottle:swing}. \label{fig:bottle:swing:stills}}
\end{figure}

\begin{figure*}
  \centering
  \includegraphics[width=\linewidth]{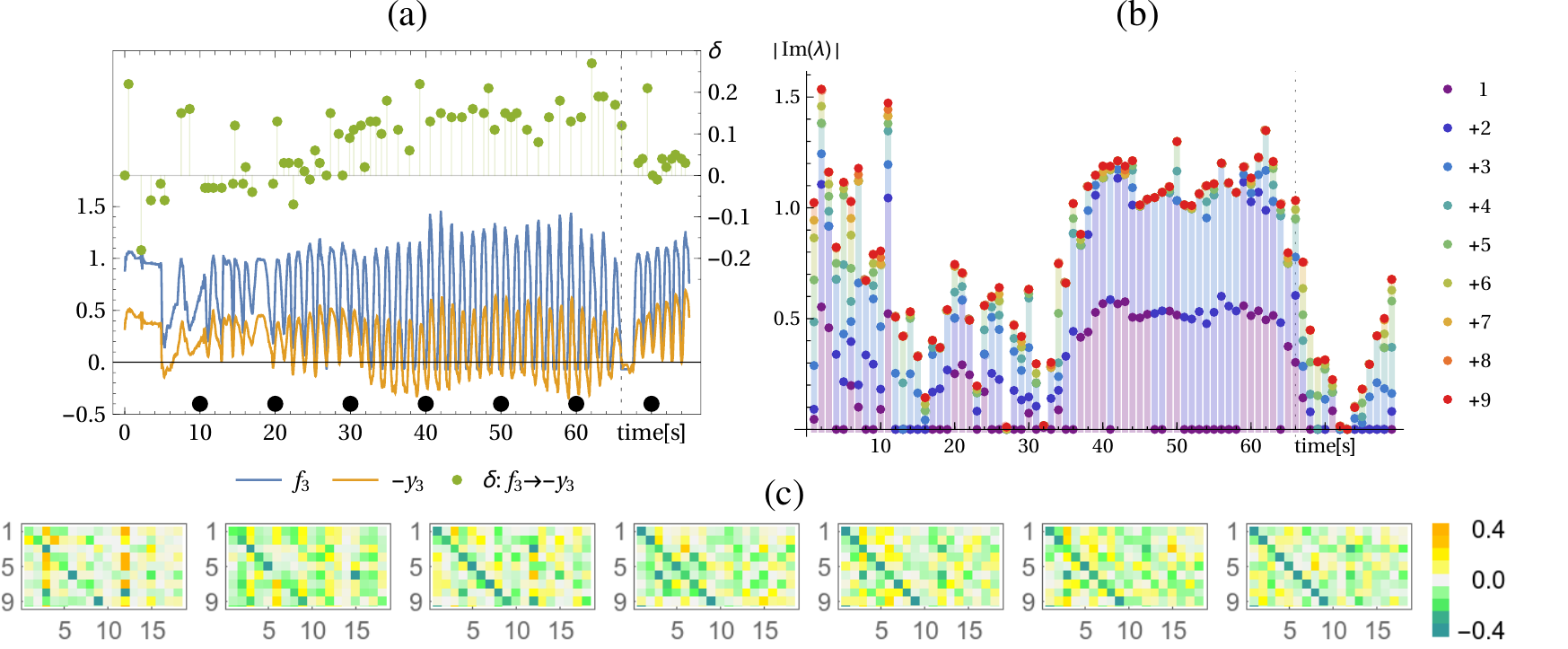}\\[-.8em]
  \caption{{\bf Bottle swing experiment.} At second $66$ the string of the bottle was shortened, see \vid{bottle:swing}.
    (a) Force sensors and control signal of muscle 3 and their time lag.
    The measured force (spring compression) and the control signal $y$ (desired tendon length) follow
    a similar trajectory with inverted sign (note $-y$). The time lag $\delta$ (right axis in seconds) between
    force and motor value (same result for other muscles) indicates that initially
    the control and the environmental influences are not in sync
    whereas in the swinging mode (from $33$\,s on) a stable phase/time-lag relation is observed.
    (b) Displayed are the absolute imaginary parts of the eigenvalues of the linearized system dynamics (Jacobian $\bar L$) (averaged over 1\,s)
    and cumulatively plotted  $(1,1+2,1+2+3,...)$. During the pronounced oscillation between 35 and 68\,sec there
    is one pair of dominant complex eigenvalues.
    (c) Corresponding controller parameter $C$ at the seconds $10,20,\dots,70$ (from left to right) as indicated by the black dots in (a).
    \label{fig:bottle:swing:analysis}}
\end{figure*}

In \vid{bottle:swing} it
can be seen how latent velocity correlations are being amplified to end up in stable circular motion patterns of the weight.
The experiment starts in a situation where the motor activities have settled
to rest, interrupted by occasional bursts so that the bottle is excited to some minor
pendulum motion. The thereby induced elongations of the springs induce the said
self-amplification of latent pendulum modes as observed in the
experiments, see also \fig{fig:bottle:swing:stills}.
These findings corroborate our claim that the subsystem---by its internal dynamics---is piloting the
meta-system  into a resonant state, \ie a whole-system mode with defined frequency.
This is also supported by the data recorded from the experiment.
When analyzing the time lag between measured force and
  driving signal (motor commands), see \fig{fig:bottle:swing:analysis}(a),
it becomes evident that initially bottle and arm are not in a fixed phase relation.
During that interval,
the structure of the controller matrix $C$ changes strongly until the system reaches the rotational mode,
see \fig{fig:bottle:swing:analysis}(c).
This illustrates the dynamics of the controller to find a self-consistent behavior.
Without perturbation this oscillatory behavior may become stationary.
However, later in the experiment the string of the pendulum was shortened such
 that a different sensorimotor coordination forms.
An indicator that the oscillations are produced by the controller is found in the complex eigenvalues
 of the linearized internal dynamics $\bar L$ as displayed in \fig{fig:bottle:swing:analysis}(b).
During the swinging mode, we find only 1 pair of significantly non-zero complex eigenvalues
 representing the main oscillatory component.
This confirms our observations that low-dimensional behaviors emerge.


\subsubsection{The emerging cocktail shaker}
\label{sec:bottle:shaker}
In a next series of experiments we attached a bottle filled with some water to the tip of the arm
in either horizontal or vertical orientation.
These experiments are meant to support our hypothesis that the meta-system may become resonant with the
internal dynamics of a subsystem, if the latter provides correlations over space and time.
This internal dynamics is making itself felt
 by the inertial forces when the water is hitting either the walls or top and bottom of the bottle.
These impacts cause a reaction of the springs and hence of the sensor values. This may
increase correlations in $\bar L$, which enhances motions of the arm in coherence with these signals.

This hypothesis is difficult to analyze by the data alone, but we can
get an impression by changing the physics of the subsystem.
In a first step, see \vid{bottle:shaking:horizontal:white},
the bottle is filled only with a little water so that the reactions of the subsystem are weak.
Nevertheless, the meta-system is already reacting to the motions of the water inside the bottle, but
the emerging modes are
only  metastable\footnote{We borrow the term metastable from physics, meaning a mode with a limited life time.}
with a very short life time. Each mode is followed by a short irregular regime until a new short-lived mode emerges through the
interplay of the arm movements and the water.

\begin{figure*}
  \centering
  \includegraphics[width=\linewidth]{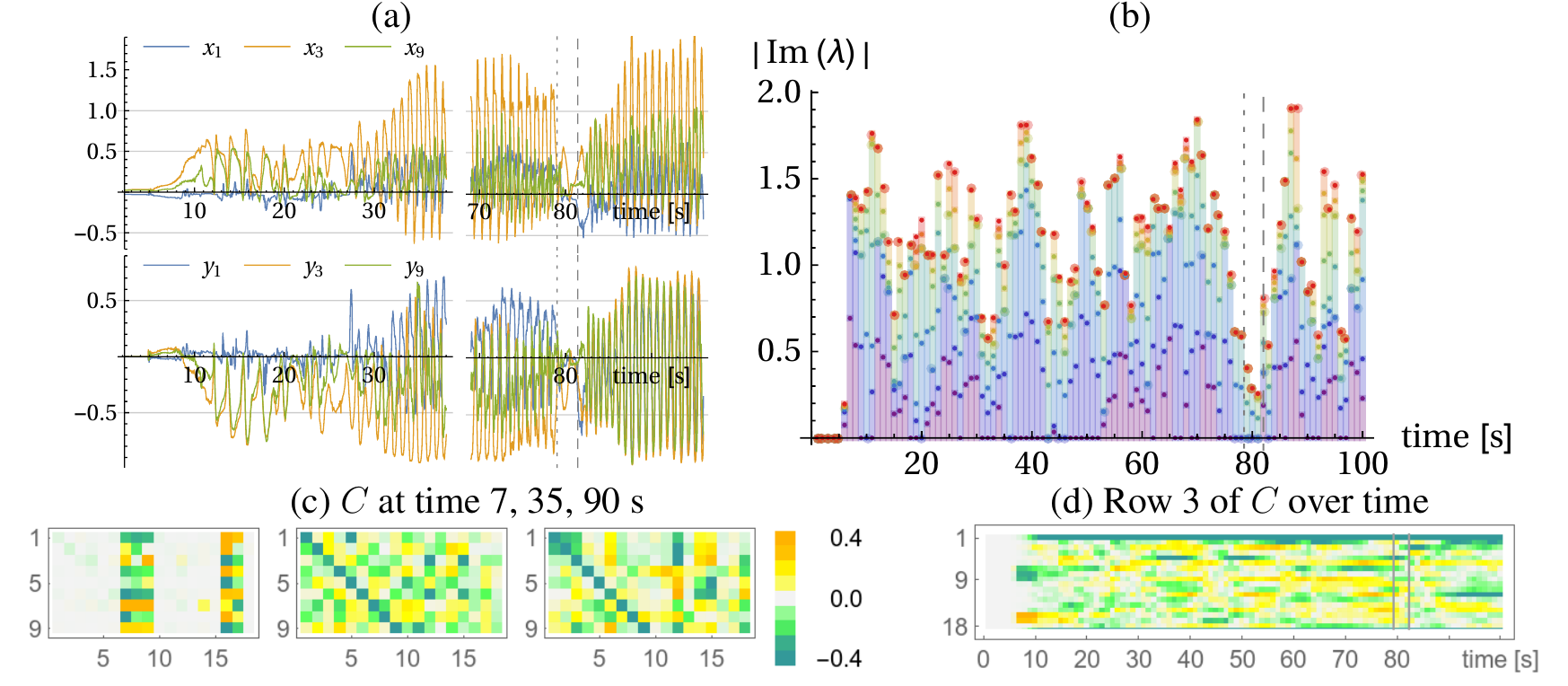}\\[-1em]
  \caption{{\bf Horizontal bottle shaking experiment.} The system starts at $C=0$ at time 0. At second $79$ the arm was
    stopped and then released at second 82 (indicated by vertical lines), see \vid{bottle:shaking:horizontal}.
    (a) sensor values ($x$) and control signals $y$ of muscle 1, 3, and 9.
    (b) Displayed are the absolute imaginary parts of the eigenvalues of the linearized system dynamics (Jacobian $\bar L$) (averaged over 1\,s)  and cumulatively plotted  $(1,1+2,1+2+3,...)$. We find one  pair of dominant complex eigenvalues accompanied by a second less strong pair.
    (c) Corresponding controller parameter $C$ at the seconds $7,25,\dots,90$ (from left to right).
    (d) The evolution of the coupling to one muscle (3) over time (corresponding to row 3 in $C$).
    \label{fig:bartender:analysis}}
\end{figure*}

In the next step, the bottle is  half-filled so that the reactive forces are much stronger. In both \vid{bottle:shaking:horizontal}
and \vid{bottle:shaking:vertical}
the modes are more pronounced and live much longer.
In \fig{fig:bartender:analysis} we present the time series data for the horizontal case.
One can see how from the initial resting position the arm starts to move until, after about 25 seconds,
 it reaches  an oscillatory mode involving strong sensor reactions (increased by the acting forces).
Interestingly, the modes are not only realized by the arm and the
attached bottle but involve also the reactive motions of the  post the arm is mounted on (see video).
The latter point can be seen if the post is fixed by hand in order to stop its swaying motion. Then, in both videos,
it can be seen that the mode breaks down immediately. Obviously, the mode involves both physical objects the
arm-shoulder system is attached to. This is not a surprise, as this system, by its dynamics,
exerts forces on both the bottle and the post leading to correlated changes in spring tensions and sensor values.
This effect is also seen in \fig{fig:bartender:analysis}(b,d) where shortly after second 40 the eigenvalues change
 and the C matrix changes until it reaches about the same configuration at second 50.
As seen already at the handshake experiment, the arm can always by manually stopped and the behavior can be changed.
At second 79 the arm was stopped and brought to rest. After the release 3 seconds later a different mode appears,
 \cf the changed motor pattern in \fig{fig:bartender:analysis}(a) and the different $C$ matrix in~(c).
To conclude, the dynamics of the $C$ matrix \eqnp{eqn:controls} is typically on a long transient (\fig{fig:bartender:analysis}(d))
 until it hits a self-consistent behavior.
Any perturbation or change in conditions leads to an adjustment of the controller and the behavior
 always aiming for a mode where high velocity correlations appear.

\subsubsection{Discovering object affordances and tool use}
The observed resonance phenomena have a close relationship to the concept of object affordances.
As introduced by \citet{Gibson1977:Affordances},
affordances are action possibilities available in the environment
to an individual. Depending on its action capabilities,
affordances define the relation between an agent and its environment through its motor and sensing capabilities
(e.g., graspable, movable, or eatable).
In this sense, in the same way as a chair affords sitting or a knob affords twisting,  we may say that
the meta-system discovers the possibility to drive the attached pendulum into  a circular motion by moving the pivot in
a definite way. This is one affordance a pendulum can offer in that specific setting.
Interestingly,  this discovery can only be made by the emerging whole-system activity with the
pendulum on the basis of the very weak interactions in the arm-pendulum physical system.
In the same sense, when giving the robot the  half-filled bottle, through the response (by the
inertia forces) of the water to the taken actions, the meta-system discovers shaking the bottle as one
possible affordance. We expect that with different objects the meta-system may discover further affordances
like the turning of a wheel as already observed  in a simulated system, see \citet{DerMartius2015:DEP}.

Object affordances may form pre-requisites for prediction and planning and are important steps in
the emergence of tool use up to generalization, abstraction,  and even creativity in cognitive robots.
We hope to support these ideas with our next experiments with the arm-shoulder system.

\subsection{Reaching}
\label{sec:reaching}
The functionality of the present control paradigm essentially rests on two different scenarios. On the one hand we have
seen a strong affinity for periodic motions like the pendulum or the bottle shaking modes.
These are predominantly observed with time scales, defined by $\rho$,  of the expected periodicity.
There, we observe that the dynamical operator $\bar L$ is essentially constant and is qualified by
(a pair of) complex eigenvalues determining the frequency of the oscillations.

With other time horizons, either shorter or not matching the period of a latent mode,
we observed  another amazing variety of behavioral patterns which give
the impression as if  the robot deliberately generates sequences of  different poses, see \vid{free:moving}.
This scenario is characterized by a $\bar L$ that is strongly varying on the time scale of
the behavior itself.
Though these sequences of movements look rather random, it is important to note that there is no artificial
randomness implanted in the controller. Instead, the controller is an explicit  deterministic function of recent sensor values.
So, we have a pseudo-random behavior
with a strong bias caused by the physical properties of the meta-system.

Noteworthy, despite their pseudo-random character, these patterns are fully
controlled as they are determined by their recent history in sensor space.
So, a pattern can be reproduced by starting the system with the history recorded in an earlier experiment.
In this sense,  motion patterns
are reproducible and may be used in higher level functional architectures.
In particular, this is of interest for generating
behavior architectures of the arm
like realizing a predefined trajectory or reaching toward a specific point.
We strongly believe that this property can open new ways for controlling such complex systems,
bypassing the need for modeling and concrete programming the behavior of the robot following the
common paradigms of control.

\section{Summary and outlook}

This paper discusses a novel approach for controlling systems with strong embodiment effects.
Instead of using specific goals or targets we aim at behavioral self-exploration
 of the system under control. Desired behaviors can then be generated on top of such basic
 behavioral primitives.
This is in contrast to the typical control approach where specific goals have to be met in an optimal fashion.
For elastically actuated robotic systems, however, we are lacking analytical models and often have no direct way to specify
 a desired behavior, except as reward and punishment signal in reinforcement learning but this
 takes prohibitively long in the systems considered here.
Our approach creates behavior only based on the responses of the system in the most recent past in an online
 fashion.
The essential new feature is the irreducible unity of the controller and the controlled
as formulated by the pair of \eqns{eqn:forward}{eqn:dotxdotz}.
The controller is devoid
of any system-related functionalities as it is given by a fixed, explicit and context-free
function of the recent history of the sensor values \eqnp{eqn:barL}.
In this interplay between the controller and the controlled, fuelled by the positive feedback strength,
the controller manages to identify and amplify tiny responses from the
world outside itself.
In our case study with the tendon-driven arm shoulder system,
if the arm alone will realize a seemingly pseudo-random but fully controlled  sequence of poses.
But more interestingly, if the arm is extended by attaching objects with an internal dynamics of its own,
the controller gets in a functional resonance with the object's internal dynamics:
metaphorically speaking, with a half-filled bottle the system develops into a cocktail
shaker and after attaching a pendulum to the tip of the arm, the controller, mediated by the arm,
drives the pendulum into a circular mode.

All these patterns emerge with great ease and in a natural and elegant way starting with $C=0$ as initial condition.
The novelty of this approach can be expressed by  the simple argument
that  comparable  motion patterns can only be generated with existing controller paradigms
 if a great amount of knowledge about the system is available or after very long training times
with handcrafted reward schemes.
Remember that the controller receives nothing but the sensor signals---the sum of tendon  plus spring length,
so the result is highly non-trivial.
Moreover, we have also given perspectives of how to extend this controller paradigm to goal oriented behavior by just
superimposing additional motor signals on the self-regulated meta-system in its ``plastic aggregation state''.
This promises a qualitative leap in controlling such robotic systems.

As a perspective, the observed response of the system to the world's internal degrees of freedom---as demonstrated with the pendulum, \eg---leads the way to
an important generalization: equipping the robot with additional sensors reporting the spatial relation of its
mechanical  degrees of freedom to the structure of the environment, we expect a similar integration of those
relations into the emerging behavioral modes. In a preliminary experiment, by integrating a camera this mechanism even lead to an active exploration of visuomotor coordination.

Finally, let us discuss on which platforms our controller is likely to create useful behavior in some way or the other. First of all, the system has to provide sensory feedback about acting physical forces to make
 embodiment effects perceivable by the controller. This is, for instance, not the
 case if all perturbations are perfectly compensated by a low-level PID controller.
Secondly, there should be sensors reporting a similar quantity as used to control the actuators, \eg
 position sensor for position control or force sensors for force control. Additional sensors are typically integrated into the loop if they show a definite response (correlation) to the motor patterns.
Thirdly, the behaviors of interest should be oscillatory.
Since we only need the main sensor-to-motor wiring information about the particular robot (which can also be learned) and do not any other specific information, we expect our system to work with a wide variety of machines
 including soft robots, but this remains for future research.

\section{Appendix: Methods}

In this section we  derive the controller matrix $C$, discuss the role of the sensor to motor mapping matrix $M$,
and give the update rule for $C$.

\subsection{The controller}
\label{sec:Appcontroller}
In order to derive \eqn{eqn:controls}, we need to predict $\dot x_t'=\dot x_{t+1}$
on the basis of the previous sensor values.
Our idea is based on an operator~\footnote{Actually, $L$ is simply a matrix but we call it an operator for emphasizing
the dependence on the states it is operating on.}
 \begin{equation}
   L_t=\dot x_t' \hat x_t^{\T}
 \end{equation}
where $\hat x = \|\dot x\|^{-2}\dot x$, which describes the transition $t\rightarrow t+1$ as
 \begin{equation}
    \dot x_{t}' = L_t \dot x_t
    \label{eqn:App:Ldyn0}
 \end{equation}
 As to notation, we remark that with any two column vectors $a$ and $b$, $a^{\T}b$ is the scalar product and
$S=ab^{\T}$ is a matrix with elements $S_{ij}=a_ib_j$. In particular, $\dot  x^{\T}\dot x=\|\dot x\|^{2}$
and $\hat  x^{\T}\dot x=1$, which corroborates \eqn{eqn:App:Ldyn0}.

 Of course, $L_t$ cannot be used directly as it involves the future.
The idea is to generalize that expression in a way which is consistent for trajectories with some time coherence.
Here, we replace $L$ with its moving average, defined as
\begin{equation}
\bar L_t = \frac 1 Z  \sum_{s=1}^{t-1} \rho^{t-1-s} L_s =  \sum_{s=1}^{t-1} p_s \dot x_{s+1} \dot x_{s}^{\T}
\label{eqn:App:barLm}
\end{equation}
where
$\rho<1$ defines the time scale for the extension of the past and
 $Z=\sum_{s=1}^{t-1} \rho^{t-1-s}$ and $p_s = \frac 1 Z \rho^{t-1-s}\| \dot x_{s} \|^{-2} $ are
 normalization and weighting factors, respectively,
and we used $\dot x_{s}'=\dot x_{s+1}$ in the last step.
Writing $x_i(s)$ for the i-th component of the vector $x_s$ and similarly for $\bar L_t$, the matrix elements of $\bar L_t$ are
\begin{equation}
\bar L_{ij}(t)  = \sum_{s=1}^{t-1} p_s \dot x_{i}(s+1) \dot x_j(s)
\label{eqn:App:barL}
\end{equation}
and in shorthand $\bar L_{ij}(t)= \langle \dot x'_i \dot x_j  \rangle_P^{t-1}$, where $p_s=\frac 1 Z \rho^{-s}\| \dot x_{s-1} \|^{-2}$ are weighting factors. Note that $\bar L$ does not any longer involve the future as it is shifted by one step in time and that the time coherence can be
controlled by the decay term $\rho$. Remembering that $\dot x_t=x_t-x_{t-1}$ we see that $\bar L$
is given by the history of the sensor values in a definite way.

We may now use  \eqn{eqn:App:barL} as dynamical operator so that
\begin{equation}
  \dot x_t' =\bar L_t \dot x_t
  \label{eqn:App:apprdyn}
\end{equation}
generates a time series: given the history ${\dot x_{t},\dot x_{t-1},\ldots}$, we get the
future evolution of $\dot x$ by iterating \eqn{eqn:App:apprdyn}.
 Note that the history until $t$ defines the future of the time series  in a deterministic way
 so that there are as many time series as there are different histories.
 Because of this generality, we may call \eqn{eqn:App:apprdyn} with \eqn{eqn:App:barL} a template defining  a
 certain class of time series. Note there are no parameters involved, apart from the time scale of the history
 as set by $\rho$.

Using  \eqn{eqn:App:apprdyn} we may express  the
future state $\dot x'$ in terms of its history. Taking the time derivative of \eqn{eqn:dotxdotz} yields $\dot y = M \dot x'$,
assuming $M$ is constant.
 Putting $y=g(z)$ where $g$ is the squashing function,
we obtain for  the internal control variable $z$,
ignoring the nonlinearity of the squashing function
 \begin{equation}
   \dot z_t \approx C_t\dot x_t
 \end{equation}
where
\begin{equation}
C_t=  M_t\bar L_t
\label{eqn:App:Cexplicit}
\end{equation}
In a final step
we have to relate $\dot z$ to $z$ in a way that is consistent with
the postulated slowly varying nature of $\bar L$. In this paper we use the
 most simple postulate,
 \ie omit the residual time dependence of C altogether.
Using the simple relation $x_t=\sum^{t}_{s=1} \dot x_s + x_0$ and analogously for $ z_t$, we define our controller as
 \begin{equation}
   y_t =
   g( C_t (x_t-x_0)  +z_0)\,,
 \end{equation}
where $z_0$ is an overall bias that can be set arbitrarily, each $z_0$ leading to a different control strategy.
This versatility is a direct result of working with the $\dot x$.
$z_0$ may also be adapted following a heuristics like avoiding a saturation
regions of the squashing function or coping with an overall bias of the system.
$x_0$ plays the role of  a sensor bias. In this paper, $x_0=0$ as our systems are centered around $x=0$.
Otherwise it can be adapted\footnote{Actually, $x_0$ is the state at time $t=0$. However, even by very small
perturbations or the residual time dependence of $C$, the memory of the initial state is soon lost
so that we are free to choose the sensor bias.}
similarly to $z_0$. In the experiments we used the controller
\begin{equation}
   y_t=g(C_t x_t)
   \label{eqn:App:controls}
\end{equation}
throughout, where $g_i(z) = \tanh(z_i)$.

For a brief discussion
we assume that $x_t$ follows a harmonic oscillation with period $T$
and ask whether this is consistent with the explicit form of $\bar L$ as given by \eqn{eqn:App:barLm}.
Linearizing
\eqn{eqn:controls} so that  $x'=My\approx MCx=\bar Lx$, we have to consider the role of $\bar L$ as applied to $x$.
Considering  \eqn{eqn:App:barLm} together with
 $\dot x_t^{\T} x_t=\dot x_{t-T/2}^{\T}x_t=0$ and  $\dot x_{t-T/4} \propto x_t$, we
may argue~\footnote{Note that the norm of the rotating vector is constant and assume $\rho\approx 1$, \ie the weighting factors $p_s$ are independent of $s$. }
that $\bar L x_t \approx \dot x'_{t-T/4}\propto x'_t$. This crude argument which repeats for each half-period
shows that the controller is consistent with the stipulated sensor dynamics, provided the mapping $M$ can
translate this into appropriate motor commands. Note also that the time smoothing in $\bar L$ does not  mean a time smoothed dynamics
as the above argument remains valid for any frequency (below the Nyquist frequency for the given update rate).

\subsection{The role of $M$}
\label{sec:roleofM}
Central to the approach is the template dynamics \eqn{eqn:App:apprdyn}.
In general, any real dynamics will not fit into that template, \ie the true dynamics in sensor space
generated by the robot can be written as
\begin{equation}
  \dot x_t'=\bar L_t \dot x_t+ \xi_t
  \label{eqn:App:barLxi}
\end{equation}
where  $\bar L$ is given by \eqn{eqn:App:barL} in terms of the real trajectory as generated by \eqn{eqn:App:barLxi}.
Let us now hypothetically assume that $A$ represents exactly the sensor response of the arm to the motor actions,
\ie assume  $\dot x'=A\dot y$ with $A=M^{-1}$ and $\xi=0$. Then, the pair of \eqns{eqn:forward}{eqn:dotxdotz} degenerates
into a triviality so that the system  can realize any trajectory.
So, the actual point of interest is the mismatch,  represented by $\xi$, between true behavior and the template.
As a rule of thumb we postulate that the controller is able to realize a trajectory the better the smaller $\xi$.
The mismatch $\xi$ directly reflects the physical reactions of the meta-system to the motor actions. In this light,
trajectories are more stable for more systematic reactions, involving the degrees of freedom of the physical system
in a coherent manner. This is what we may observe in the experiments.

The standard version is that $M$ is to reflect just the most basic causal relations between sensor and motor signals.
$M$ can be learned in simple off-line motor babbling scenarios, see~\citet{DerMartius2015:DEP} for examples.
In this paper, we used the relation between motor encoder and tendon length which is a one-to-one mapping, hence
$M$ is the  unit matrix (identity operator). This choice also underlines the difference between the matrix $M$ and the usual understanding
of an internal inverse model: while the latter is to reflect the mapping from sensors to motors as precisely as possible,
$M$ determines $\xi$---the mismatch between template and true dynamics which determines both the self-exploration rate
and the regularity of the generated motion patterns as a (very complex) function of the character parameters ($\rho$, $\kappa$).
More details on this search and converge paradigm may be found in~\citet{DerMartius2015:DEP}.

\subsection{Eliciting periodicity}
\label{sec:periodicity}
In order to make the meta-system more attractive for motion patterns of a certain period $T$, we introduce a vector $x_{t-T/2}$ of delay
sensor values  together with a second controller matrix $C_{t|-T/2}$ and define the vector of the motor commands as
\begin{equation}
  y_t = g(C_t x_t + C_{t|-T/2}x_{t-T/2})
\end{equation}
where, before normalization,
\begin{equation}
  C_{t|-T/2}=M_{t|-T/2}\langle \dot x'_i \dot x_j  \rangle_P^{t-T/2-1}
\end{equation}
and $M_{t|-T/2} $ transforms motor values at time  $t$ into sensor values at time $t-T/2$.
Note, that $M_{t|-T/2} $ can be different than $M_t$.
If the system is in a periodic regime we have  $x_t=-x_{t-T/2}$ and thus also $\langle \dot x'_i \dot x_j  \rangle_P^{t-1}=\langle \dot x'_i \dot x_j  \rangle_P^{t-T/2-1}$.
Choosing $M_{t|-T/2}=-M_t$, we find $C_{t|-T/2}=-C_t$ so that, before the normalization, the argument of the squashing function is given by
$C_t x_t + C_{t|-T/2}x_{t-T/2}=2C_t x_t$. Hence, there is a constructive interference over the half period so that periodic patterns
are getting favored for self-amplification. This argument holds true also for any multiple of the fundamental period $T$.

\subsection{Some technical details}
\label{sec:App:technicaldetails}
In the practical applications done so far, we identified  a number of tweaks for coping with peculiarities of the approach.
One is the choice of the time lag between $x$ and $x'$ which was $x'_t=x_{t+1}$ above. However, there is no hindrance to
introduce a certain lag $\theta$ so that $x_t'=x_{t+\theta}$. This is helpful in  order to adapt the system to the actual
update rate and in particular for enhancing the chance for periodic patterns. It also influences to some extent the
 frequency of such  patterns.

Another point concerns the regularization of the normalization factors which have different effects for either the normalization of
$C$ or that of $\bar L$ which was introduced with the $p_s$ factors \eqn{eqn:App:barL}. As the former normalization acts on $C$ directly, its
effect is delayed on the time scale given by $\rho$.  The normalization of $\dot x$ on its hand
also needs some regularization, \ie we have to replace the factor $\|\dot x\|^{-2}$ as
\begin{equation}
  \|\dot  x\|^{-2} \rightarrow \frac{1}{\|\dot x\|^{2}+r}
\label{eqn:App:normdotx}
\end{equation}
where $r$ may run in principle  from $10^{-1}$ down to a minimal value determined by the discretization of the sensor values.
However, very small $\dot x$ are enlarged up to a factor $r^{-1}$
and dominate in this way the definition of $C$. This is not helpful as in most cases the very small velocities arise from \eg sensor noise,
 which tends to destroy the already reached self-amplification of latent behavior. In practice, it is therefore helpful
to keep the regularization effect in bounds. In other words, while we were using $r=10^{-10}$ for this paper,
values of $r=10^{-3}$ seem to work better in current  experiments.

\subsection{Update rule}
\label{sec:Appupdate}

We can also give an update rule for $C$ as
\begin{equation}
   \tau \dot C_t = M_t  \dot x'_t \hat x_t^{\T} - C_t
 \end{equation}
using  $\hat x = \dot x \|\dot x\|^{-2}$ or its regularized version given by \eqn{eqn:App:normdotx}.
With discrete time, we write the update of $C$ as
\begin{equation}
 \tau \Delta C_t= M_t\dot x'_{t} \hat x_{t}^{\T} -C_t
\end{equation}
which becomes stationary if $C=M\bar L$.

\section{Acknowledgement}
We thank Alois Knoll for inviting us to work with the Myorobotic arm-shoulder system at the TUM.
Special thanks goes also to Rafael Hostettler
 for helping us with the robot and control framework.
 RD thanks for the hospitality at the Max-Planck-Institute and for helpful discussions with Nihat Ay and Keyan Zahedi.
GM received funding from the People Programme (Marie Curie Actions) of the European Union's Seventh Framework Programme (FP7/2007-2013) under REA grant agreement no.~[291734].


\end{document}